\pdfoutput=1

\documentclass[11pt]{article}

\usepackage{EMNLP2022}

\usepackage{times}
\usepackage{latexsym}

\usepackage[T1]{fontenc}

\usepackage[utf8]{inputenc}

\usepackage{microtype}

\usepackage{inconsolata}

\title{Does Self-Rationalization Improve Robustness to Spurious Correlations?}

  \author{\begin{tabular}{c}
    Alexis Ross$^{\dagger*}$ \quad
    Matthew E. Peters$^\ddag$ \quad 
	Ana Marasovi\'{c}$^{\S*}$
	\end{tabular}
	\\ \vspace{.5mm}
	\begin{tabular}{c}
	\\ \vspace{.5mm}
	$^\dagger$Massachusetts Institute of Technology, Cambridge, MA, USA \\
	$^\ddag$Allen Institute for AI, Seattle, WA, USA \\
	$^\S$University of Utah, Salt Lake City, UT, USA \\
	\end{tabular}
	\\ \vspace{.5mm}
    \begin{tabular}{c}
\texttt{alexisro@mit.edu} \quad  \texttt{matthewp@allenai.org} \quad \texttt{ana.marasovic@utah.edu}
\end{tabular}
}
\vspace{5mm}

\usepackage{graphicx}
\usepackage{xcolor}
\usepackage{colortbl}
\usepackage{color}
\usepackage{xspace}
\usepackage{array}
\usepackage{tabulary,booktabs}
\usepackage{comment}
\usepackage{multirow}
\usepackage{makecell}

\newcommand{\highlight}[1]{\note{blue}{#1}}
\newcommand{\newhighlight}[1]{\note{purple}{#1}}
\renewcommand{\highlight}[1]{{#1}}
\renewcommand{\newhighlight}[1]{{#1}}

\newcommand{\cqa}{\textsc{Cqa}\xspace}
\newcommand{\ecqa}{\textsc{Ecqa}\xspace}

\newcommand{\nli}{\textsc{Nli}\xspace}
\newcommand{\snli}{\textsc{Snli}\xspace}
\newcommand{\esnli}{\textsc{eSnli}\xspace}
\newcommand{\tfive}{\textsc{T5}\xspace}
\newcommand{\bart}{\textsc{Bart}\xspace}
\newcommand{\gpt}{\textsc{Gpt2}\xspace}
\newcommand{\hans}{\textsc{HANS}\xspace}
\newcommand{\cad}{\textsc{CAD}\xspace}
\newcommand{\dev}{\textsc{Test}\xspace}

\newcommand{\gptmedium}{\textsc{Gpt2-Medium}\xspace}
\newcommand{\gptlarge}{\textsc{Gpt2-Large}\xspace}

\newcommand{\bartbase}{\textsc{Bart-Base}\xspace}
\newcommand{\bartlarge}{\textsc{Bart-Large}\xspace}

\newcommand{\tfivelarge}{\textsc{T5-Large}\xspace}
\newcommand{\tfivebase}{\textsc{T5-Base}\xspace}

\newcommand{\snlihard}{\textsc{Test-Hyp}\xspace}

\newcommand{\devhard}{\textsc{Test-Hard}\xspace}
\newcommand{\deveasy}{\textsc{Test-Easy}\xspace}

\newcommand{\minus}{\scalebox{0.75}[1.0]{$-$}}

\newcommand{\deltaspread}{$\Delta$ \textsc{Test-Subsets}\xspace}

\newcommand{\spread}{\textsc{Test-Easy}~\minus~\textsc{Test-Hard}}

\newcommand{\ie}{\emph{i.e.,}\xspace}
\newcommand{\eg}{\emph{e.g.,}\xspace}

\definecolor{lightpurple}{RGB}{203, 195, 227}
\definecolor{lightblue}{RGB}{176,224,230}

\newcommand\sect[1]{\S\ref{#1}}

\newcommand\blfootnote[1]{%
	\begingroup
	\renewcommand\thefootnote{}\footnote{#1}%
	\addtocounter{footnote}{-1}%
	\endgroup
}

\begin{document}
\maketitle

\begingroup\def\thefootnote{*}\footnotetext{Work undertaken while Alexis Ross and  Ana Marasovi\'{c} were at the Allen Institute for AI.}\endgroup

\blfootnote{Our code is publicly available at \url{https://github.com/allenai/rationale_robustness}}
\begin{abstract}
Rationalization is fundamental to human reasoning and learning. NLP models trained to produce rationales along with predictions, called self-rationalization models, have been investigated for their interpretability and utility to end-users. However, the extent to which training with human-written rationales facilitates {learning} remains an under-explored question. We ask whether training models to
self-rationalize can aid in their learning to solve tasks \textit{for the right reasons}. 
{Specifically, we evaluate how \highlight{training self-rationalization models with free-text rationales}
affects robustness to spurious correlations in fine-tuned encoder-decoder and decoder-only models of six different sizes}. We evaluate robustness to spurious correlations by measuring performance on 1) manually annotated challenge datasets and 2) subsets of original test sets where reliance on spurious correlations would fail to produce correct answers. {We find that while self-rationalization can improve robustness to spurious correlations in low-resource settings, it tends to hurt robustness in higher-resource settings. Furthermore, 
these effects depend on model family and size, as well as on rationale content.} Together, our results suggest that explainability can come at the cost of robustness; thus, appropriate care should be taken when training self-rationalizing models with the goal of creating more trustworthy models. 
\end{abstract}
\section{Introduction}

{Rationalization---the process of explaining the reasoning used to come to a particular decision---plays a pivotal role in human inference and learning \citep{explanation-lombrozo}.} 
For these reasons, there has been a growing interest in producing NLP models that can output rationales\footnote{{Prior work has used the terms ``explanation'' and ''rationale'' interchangeably. In this work, we use the word ''rationale'' for consistency with ''self-rationalization'' models.}} for their predictions. Models that output such rationales have multiple benefits: First, they are more interpretable and easier to interact with for end-users than non-rationalizing models \citep{melis-self-explaining-robust}. Second, such intermediate rationalization can offer learning benefits, such as achieving comparable performance with less data and improving out-of-distribution generalization \citep{scratchpads, chain-of-thought, star}.

However, the question of whether training models to rationalize can help them learn how to solve tasks \textit{for the right reasons} remains open. In particular, rationales encode information about the underlying reasoning humans use to reach answers, which raises the question: Does incorporating such rationales into training allow models to rely on human-aligned reasoning rather than spurious feature interactions? {If so, training with rationales could offer a pathway towards creating more robust, trustworthy, or cognitively plausible models.}

In this work, we explore this question by empirically investigating whether training models with human-written rationales can help make them more robust to spurious correlations in data. We analyze a class of models called self-rationalization models---which jointly output free-text rationales along with predictions---and focus specifically on the fine-tuning setting, in which prior work has found reliance on spurious correlations to emerge \citep{utama-etal-2021-avoiding}.

We evaluate six models of varying architectures and sizes across two tasks, natural language inference and commonsense question answering. Our main results are as follows: \begin{enumerate}
    \item While the effects of training with rationales are model- and task-specific, 
    when it improves robustness to spurious correlations, it tends to be in lower-resource settings. In higher-resource settings, training with rationales can hurt robustness (\sect{ssec:overall}). 
    \item Within model families, larger models benefit more in robustness from rationales (\sect{ssec:model_size}).
    \item {The effects of self-rationalization on robustness are not fully explained by
    its effects on in-domain task performance} (\sect{ssec:correlation}).
    \item The content of rationales used during training influences both task performance and robustness to spurious correlations (\sect{ssec:rationale_content}).
\end{enumerate} 

Our results suggest that straightforward self-rationalization training does not always facilitate learning to solve a task for the right reasons. Instead, the effects of self-rationalization on robustness to spurious correlations depend on a multitude of factors. Thus, appropriate care should be taken when training models to self-rationalize for the goal of creating trustworthy models.

\section{Related Work}
\paragraph{Learning to rationalize} 
Two classes of approaches to producing models that can rationalize their predictions include \emph{self-rationalization} models,\footnote{\highlight{Such approaches have also been referred to as \emph{explain-then-predict} \citep{esnli} and \emph{rationalize-then-predict} \citep{rationalization-robustness} models.}} which are fully differentiable and output free-text rationales along with task predictions, and \emph{pipeline} models, which consist of two components---one that produces rationales, and a second that makes predictions from those rationales \citep{wiegreffe-etal-2021-measuring}.\footnote{\highlight{See \citet{wiegreffe-etal-2021-measuring} for a detailed discussion of pipeline and self-rationalization approaches to rationalization.}} 
\highlight{Such methods are typically evaluated by the \emph{faithfulness} and \emph{plausibility} of their rationales, where faithfulness represents the extent to which a model actually relied on the rationale in making its prediction, and {plausibility} indicates human judgment of how well the rationale explains the output
\cite{deyoung-etal-2020-eraser}.}

\highlight{In contrast to these works which aim to improve model interpretability through new methods for rationalizing models,
we ask to what extent existing methods affect model robustness to spurious correlations.
We conduct our analysis on self-rationalization models, which have been found to achieve better task performance and produce higher-quality rationales than do pipeline models \citep{wiegreffe-etal-2021-measuring,esnli}.}

\vspace{-0.25mm}

\paragraph{Learning from rationales} Recent work has explored the utility of rationales for improving end-task performance in in-context learning \citep{chain-of-thought, deepmind-prompt, unreliability-few-shot} as well as in fine-tuning \citep{zaidan-etal-2007-using,hancock-etal-2018-training,esnli, wt5, expl-formal-framework, scratchpads, lirex}.
Previous work has shown that training with both human-annotated rationales \citep{rajani-etal-2019-explain} and rationales generated by language models \citep{paranjape-etal-2021-prompting} can increase in-domain task performance, particularly in low-resource settings \citep{bhat-etal-2021-self, teacher-student-expl, star}.
{Unlike these prior works, which study how training with rationales affects in-domain, end-task performance, we focus specifically on evaluating impact on robustness to spurious correlations.}

\vspace{-0.25mm}

\paragraph{Improving robustness with rationales} Most closely related are recent works that study how training with rationales affects model robustness.
\citet{Stacey2021NaturalLI} propose a method of supervising attention weights with extractive rationales and show that this method leads to both in-distribution and out-of-distribution improvements for natural language inference.
\citet{schuster-etal-2021-get} find that training with contrastive extractive rationales improves robustness as measured by performance on adversarial evaluation sets. 
Concurrent work by \citet{rationalization-robustness} investigates to what extent training models to extract rationales through pipelines improves their robustness to adversarial attacks.

In contrast to all three of these works, we focus on freeform rationales instead of extractive rationales and explore the impact of amount of training data on robustness.
In contrast to \citet{schuster-etal-2021-get} and \citet{rationalization-robustness}, we analyze self-rationalization models instead of pipeline models and measure robustness to spurious correlations, rather than robustness to adversarial attacks. While \citet{Stacey2021NaturalLI} evaluate robustness to spurious correlations for natural language inference with some of the same test sets, they work with masked language models and evaluate the effect of supervising model attention with rationales; in contrast, we work with encoder-decoder and decoder-only models of varying sizes and evaluate the effect of outputting rationales along with predictions. In addition, their analysis is limited to natural language inference, for which evaluation datasets targeting robustness exist; in contrast, we also experiment with commonsense question answering through \newhighlight{new methods for evaluating robustness. In \sect{ssec:overall}, we discuss the variance in results across different tasks and highlight the importance of cross-task evaluation.}
\section{Experiments}

\subsection{Experimental Set-Up}

\paragraph{Models} 
We experiment with encoder-decoder and decoder-only models of varying sizes {ranging from 140 to 774 million parameters}, as shown in Figures~\ref{fig:nli-freeform} and \ref{fig:cqa-freeform}. 
Our encoder-decoder models build on pretrained \textbf{\tfive} \citep{2020t5} and \textbf{\bart} models \citep{lewis-etal-2020-bart}, and our decoder-only models build on pretrained \textbf{\gpt} \citep{gpt} models. 
Our \tfive models build specifically on the versions trained for an additional 100K steps on the language modeling objective after pretraining \citep{lester-etal-2021-power}, as we aim to measure how the amount of training data impacts results, and the default \tfive models have already been fine-tuned on the full \snli training dataset.\footnote{{For example, when experimenting with \tfivebase, we work specifically with \texttt{t5-base-lm-adapt} available in \texttt{huggingface} at \url{https://huggingface.co/google/t5-base-lm-adapt}.}}

\paragraph{Tasks} We evaluate self-rationalization models on two tasks---\textbf{natural language inference} (\nli), and \textbf{commonsense question answering} (\cqa)---{for which human-annotated rationales already exist}.
For \nli, we train task models on \snli \citep{bowman-etal-2015-large} and obtain rationales from \esnli \citep{esnli}. {For \cqa, we train task models on \cqa \citep{talmor-etal-2019-commonsenseqa} and obtain rationales from \ecqa \citep{ecqa}. Examples of inputs and outputs for both tasks are shown in Table~\ref{tab:input-output}.
For \cqa, unless otherwise specified, we train on the ``positive'' freeform rationales in \ecqa, which explain why the gold answer is the correct answer for a given question. In \sect{ssec:rationale_content}, we explore the impact of training with the different forms of rationales shown in Table~\ref{tab:input-output}.

\paragraph{Rationales} For each task, we compare a baseline model trained solely to predict task labels with models trained to also self-rationalize. All self-rationalization models are trained to generate a rationale following the task label, as previous work has found that outputting rationales conditioned on labels leads to better performance than outputting labels conditioned on rationales {in the fine-tuning setting}  \citep{schuff-etal-2021-external}.

\paragraph{Data} We experiment with different numbers of training examples $n$, {as we seek to understood how training data size influences the impact of self-rationalization training on robustness to spurious correlations.}
We experiment with $n \in$ \{1K, 2.5K, 5K, 10K, 50K, 100K\} for \nli and $n \in$ \{1K, 5K, 7598\} for \cqa.\footnote{The total size of original training datasets are 549,339 for \snli and 7,598 for \cqa.} For each training data amount $n$, we create validation data for checkpointing models by randomly sampling $n/2$ instances from the original \emph{task-only} validation dataset, such that we perform model selection based on task performance across baseline and self-rationalization models. For self-rationalization models, we create training data by concatenating original task-only training input-output pairs with their rationale-extended counterparts, such that we have $2n$ training inputs obtained from $n$ original instances.\footnote{In initial experiments, we find that this leads to better performance/robustness measures than only using the $n$ input-outputs for self-rationalization; we hypothesize that without including the original task-only inputs as well, self-rationalization models may be overfitting to the rationale generation part of the training objective.}

\paragraph{Training}
For each amount of training data $n$, we report the average difference between task-only and self-rationalization models across multiple random seeds (5 for \nli and 10 for \cqa).\footnote{{We experiment with more seeds for \cqa because we have fewer metrics/evaluation datasets to measure robustness for \cqa, and so it is harder to disentangle real effects from noise.}}
{For one random seed in each evaluation setting (where a setting is determined by the task, model family, model size, whether rationales are used, and amount of training data)}, we tune the learning rate from possible values $[1\mathrm{e}{-5}, 3\mathrm{e}{-5}, 5\mathrm{e}{-5}]$ and use the best-performing learning rate for other random seeds in the same setting.
We train with fixed batch size 64 and linear learning rate scheduler using Adafactor until accuracy on the validation data stops decreasing, or for a maximum of 50 epochs. {We use patience values of $10$ for $n <$ 10K, $5$ for $n >=$ 10K, and $3$ for $n >=$ 50K.}

\paragraph{Evaluation} We decode predictions using greedy decoding and evaluate accuracy using exact match with gold labels. 
{We evaluate robustness to spurious correlations by measuring performance on 1) manually annotated challenge datasets and 2) subsets of original test sets where reliance on spurious correlations would fail to produce correct answers. Both methods are discussed below in \sect{ssec:evaluating-spurious}.}

\subsection{Evaluating Reliance on Spurious Features}
\label{ssec:evaluating-spurious}

\paragraph{Out-of-domain challenge datasets} Our first method of evaluating reliance on spurious correlations leverages out-of-domain evaluation sets designed by experts to test for reliance on spurious features. For \nli, we evaluate on \hans \cite{mccoy-etal-2019-right} and \cad \cite{kaushik2021learning}. \textbf{\hans} is a controlled evaluation dataset that tests for reliance on surface-level syntactic biases present in \snli. \textbf{\cad} is an evaluation dataset with human-annotated edits to inputs that change entailment labels. 
To the best of our knowledge, such evaluation datasets do not exist for \cqa.

\begin{table}[htb]
\centering
\small
\begin{tabular}{c@{\hspace{1.25\tabcolsep}}c|c@{\hspace{1.25\tabcolsep}}r@{\hspace{1\tabcolsep}}c}
\toprule
\multicolumn{2}{c|}{\normalsize \cqa} & \multicolumn{2}{c}{\normalsize \nli}\\
\textbf{Feature} & \textbf{z} & \textbf{Feature} & \multicolumn{2}{c}{\textbf{z}}\\\midrule
fountain & 3.50 & lex-overlap $\geq$ 0.8 & 140.16 & (e) \\
music & 3.18 & for & 93.86 & (n) \\
welcome & 3.01 & to & 83.30 & (n) \\
atlas & 3.00 & sleeping & 80.68 & (c) \\
satisfied & 3.00 & there & 78.11 & (e) \\
hard & 2.98 & outside & 77.64 & (e) \\
stage & 2.86 & nobody & 68.44 & (c) \\
tale & 2.86 & outdoors & 65.17 & (e) \\
amusement & 2.65 & no & 52.58 & (c) \\
feel & 2.65 & cat & 50.72 & (c) \\
\bottomrule
\end{tabular}
\caption{Features found to have statistically significant correlations with labels in training datasets using \citet{gardner-etal-2021-competency}'s framework. 
Details in \sect{ssec:evaluating-spurious}.
}
\label{tab:artifacts}
\end{table}
\begin{figure*}[htb]
    \centering
    \includegraphics[height=0.44\textheight]{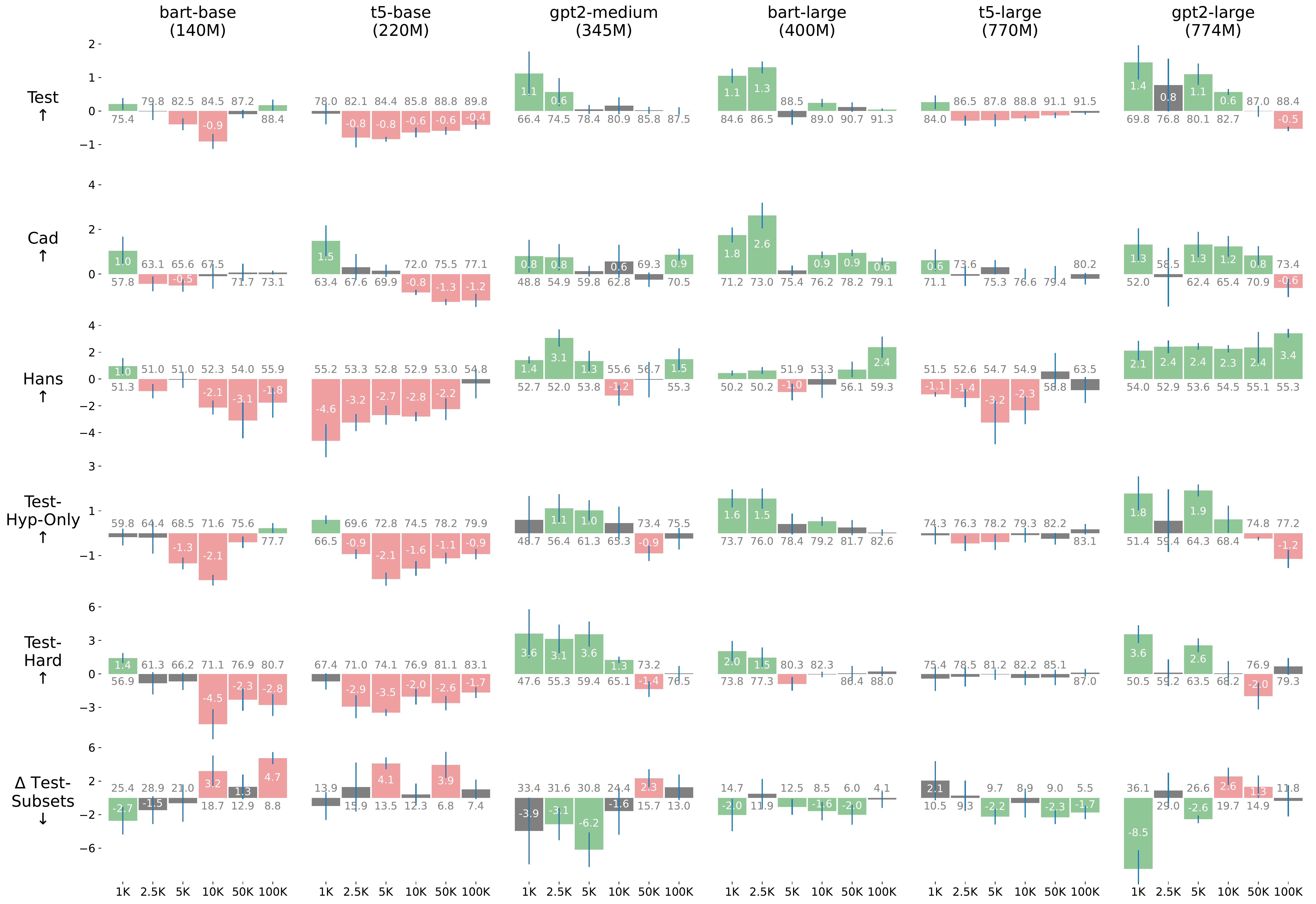}
\caption{Effect of self-rationalization for \nli across six models (columns) and varying amounts of training data (x axis). Bar heights show mean differences between baseline task-only models and self-rationalization models for: task performance (row 1), performance on manually annotated challenge datasets (rows 2-3), performance on hard subsets of original \snli evaluation data (rows 4-5), and \deltaspread (row 6). Baseline values are shown in gray. Error bars indicate standard errors of the means. Green/red bars indicate improvement/degradation in robustness; gray bars indicate error bars intersecting 0.}

\label{fig:nli-freeform}
\end{figure*}

\paragraph{``Hard'' subsets of original evaluation data}
To directly test for reliance on spurious correlations without introducing additional domain shifts, we also subset the original task test sets into subsets of varying difficulty, where {difficulty is measured by the success of spurious heuristics: ``Easy'' subsets include instances for which heuristics that build on spurious correlations in training data would lead to correct predictions, and ``hard'' subsets include instances where such spurious heuristics would fail.}

To create these ``easy'' and ``hard'' subsets,
we build on the statistical framework for uncovering dataset-level artifacts
introduced by  \citet{gardner-etal-2021-competency}. Specifically, we measure correlation between features and outputs across the \cqa and \snli training datasets and consider as artifacts any features showing statistically significant correlation, \ie with z-statistic $>$ 2.

For \snli, we consider tokens in inputs as features, as well as lexical overlap between premise and hypothesis. Following previous work \citep{wu-etal-2022-generating}, we consider an input to have high lexical overlap if the ratio of tokens in the hypothesis that are also present in the premise is at least 0.8. We use classification labels as outputs.
For \cqa, {the feature and output spaces are less clearly defined, as it contains different output choices for each input.} We take tokens in answer choices to be features and whether or not those tokens are present in the gold answers as outputs.
{To remove features that are very frequent or infrequent,} we filter features that appear less than 10 or more than 200K times for \snli and less than 5 or more than 10K times for \cqa.
Table~\ref{tab:artifacts} displays the 10 features with highest z-statistics for the \cqa and \snli training sets.\footnote{We note that the z-statistics for \snli artifacts are much higher than for \cqa; this finding aligns with prior work showing that \snli contains many artifacts \citep{poliak-etal-2018-hypothesis,gururangan-etal-2018-annotation, wallace-etal-2019-universal}. \snli train has 3,496 total artifacts, and \cqa train has 43.} 

We subset the original \cqa and \nli test sets based on whether artifacts appear with the same output {they showed statistically significant correlations with in the training datasets.} 
\textbf{\devhard} contains instances for which relying solely on artifacts to make predictions would fail to produce correct predictions (\ie artifacts appear with a different output than they are correlated with), and \textbf{\deveasy} contains instances for which relying on artifacts would lead to correct predictions.\footnote{{If an instance $(x, y_i)$ contains \emph{both} artifact(s) that show statistically significant correlation with $y_i$ in the training data \emph{and} other artifact(s) that show statistically significant correlation with $y \neq y_i$ in the training data, we exclude this instance from both \deveasy and \devhard.}} For example, a \cqa test instance for which an incorrect answer choice had token ``fountain'' would be considered ``hard,'' as ``fountain'' has statistically significant correlation with being in the correct answer choice (Table~\ref{tab:artifacts}). The sizes of \deveasy and \devhard are 76/333 respectively for \nli and 82/372 for \cqa. 
In addition to reporting performance values for these subsets, we measure the \textit{spread} in performance on hard vs. easy subsets, \ie \spread, which we refer to as \textbf{\deltaspread}. We take a lower value of \deltaspread to indicate less reliance on artifacts.\footnote{{We observe that \deveasy and \devhard in fact have the expected difficulties as measured by accuracy values of different models. In particular, as shown in the last rows of Figures~\ref{fig:nli-freeform} and \ref{fig:cqa-freeform}, the large \deltaspread values, annotated in gray, indicate that baseline models perform noticeably worse on \devhard than on \deveasy for both \nli and \cqa. In addition, baseline accuracies on \devhard are notably worse than} {accuracies on the full test sets (row 1) for \nli. While this latter trend does not hold as consistently for \cqa, we observe that the baseline accuracies on original test sets are lower for \cqa than for \nli. Thus, we hypothesize that for \cqa, the relative lack of drop in performance on \devhard compared to original test sets can be explained by the fact that \ecqa contains fewer artifacts and so original test sets are already ``hard'' for \cqa models in the sense of prevalence of artifacts to be exploited.}}
For \nli, we also evaluate on \textbf{\snlihard}, a subset of the \snli test set for which a hypothesis-only classifier was found to give incorrect predictions \citep{gururangan-etal-2018-annotation}.\footnote{We do not evaluate on the analogous ``easy'' counterpart for \snlihard, \ie the subset for which a hypothesis-only classifier succeeds, as it would require re-training a hypothesis-only classifier; instead, we evaluate only on the \snlihard subset released by \citet{gururangan-etal-2018-annotation}.}

\begin{figure*}[htb]
    \centering
    \includegraphics[height=0.28\textheight]{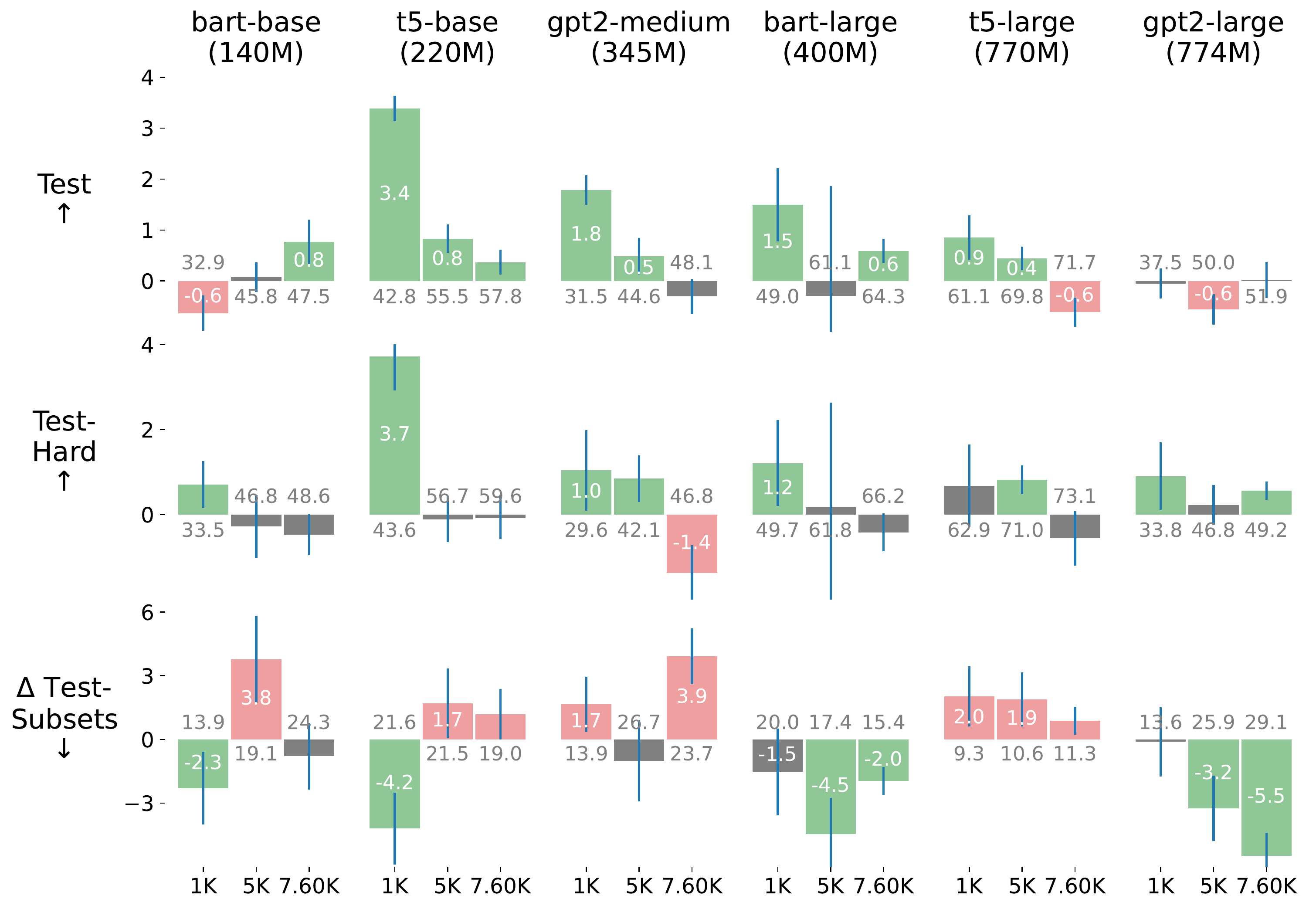}
\caption{Effect of self-rationalization for \cqa across six models (columns) and varying amounts of training data (x axis). Bar heights show mean differences between baseline task-only models and self-rationalization models for: task performance (row 1), performance on the \devhard subset of original \cqa evaluation data (rows 2-3), and \deltaspread (row 4). 
Baseline values are shown in gray. Error bars indicate standard errors of the means. Green/red bars indicate improvement/degradation in robustness; gray bars indicate error bars intersecting 0.}

\label{fig:cqa-freeform}
\end{figure*}

\begin{table*}[t]\footnotesize\centering
\begin{tabular}{l@{\hspace{0.43\tabcolsep}}l@{\hspace{0.55\tabcolsep}}|p{0.82\linewidth}}\toprule

\multicolumn{3}{p{0.97\linewidth}}{\textbf{\nli input}: 

snli hypothesis: The family is sitting down for dinner. premise: A couple play in the tide with their young son.}\\\midrule

\snli & & contradiction \\
\esnli & {}  &  contradiction explanation: The family cannot simultaneously be playing in the tide and sitting down to dinner. \\
\midrule\midrule

\multicolumn{3}{p{0.97\linewidth}}{\textbf{\cqa input}: 

Where would you get a dog if you 
do not 
have one? (A) pet store (B) outside (C) neighbor's house (D) park (E) rug}\\\midrule
{\cqa} & & pet store\\
\ecqa & {Positive} & pet store explanation: Pet store is a retail business which sells different kinds of animals, variety of animal supplies and pet accessories to the public.
We would get a dog at a pet store, if we do not have one. \\
\ecqa & {Negative} & pet store explanation: Outside means not in the living or working area. Those found outside in streets are street dogs and are not usually a pet dog.
Neighbor’s house is the man who lives in the house next door to your house is your neighbor and the house in which he lives is your neighbor’s house. The one at neighbor’s house is his pet dog and we can’t get it even though we don’t have one.
Park is a large public garden or area of land used for recreation. Dogs found at park are street dogs not suitable for being a pet.
Rug is a floor covering of thick woven material or animal skin, typically not extending over the entire floor. Rug is a floor covering and not a pet shop where we can get a dog.\\
\ecqa & {Freeflow} & pet store explanation: Pet store is a retail business which sells different kinds of animals, variety of animal supplies and pet accessories to the public. We would get a dog at a pet store if we do not have one. Those found outside in streets are street dogs and are not usually a pet dog. The one at neighbor’s house is his pet dog and we can’t get it even though we don’t have one. Dogs found at park are street dogs not suitable for being a pet. Rug is a floor covering and not a pet shop where we can get a dog. \\
\bottomrule\end{tabular}\caption{Examples of inputs and outputs used for training baseline and self-rationalization models.}\label{tab:input-output}\end{table*}

\section{Results}
\label{sec:results}

Figures~\ref{fig:nli-freeform} and \ref{fig:cqa-freeform} show, for \nli and \cqa respectively, the effects of self-rationalization across multiple random seeds.
Plotted are mean differences between self-rationalization models and baseline task-only models (\ie self-rationalization $\minus$ baseline) across six models (columns)
and varying amounts of training data (x axis). 
{Improvements on \dev (row 1) reflect in-domain, task improvements, while improvements on other metrics (rows > 1) indicate robustness improvements.}

\subsection{Main Results}
\label{ssec:overall}

As shown in Figure~\ref{fig:nli-freeform}, under our evaluation of robustness to spurious correlations, we observe that self-rationalization improves the robustness of \bart- and \gpt-based \nli models in lower resource data settings.
In higher resource settings, we observe a degradation in some robustness metrics, namely performance on \snlihard \& \devhard and \deltaspread for all models except \bartlarge. For \bartbase, this degradation in higher-resource settings is also seen for performance on \hans. 
The \tfive models (\tfivebase \& \tfivelarge) show more mixed results: While self-rationalization hurts performance on \hans for both \tfivebase and \tfivelarge in all data regimes, it improves performance on some metrics, \ie \deltaspread in higher-resource settings (n>=5k) for \tfivelarge.\footnote{One distinct property of \tfive models is that they were pretrained with a denoising objective and then adapted with a language modeling (LM) objective, while \bart was pretrained only with denoising and \gpt only with LM. Thus, we speculate that an explanation for the difference in results from the \tfive models could be that the objectives used to pretrain a model before fine-tuning may influence how self-rationalization affects robustness to spurious correlations, but why exactly the objectives may have such an effect remains unclear. }

For \cqa (Figure~\ref{fig:cqa-freeform}), results are more mixed, and they depend on model properties, \ie architecture and size, as well as size of the training data. For \bart and \gpt models of size \textsc{Large}, training with rationales generally leads to improvements. For models smaller than size \textsc{Large}, as well as both \tfive models, the effect of training with rationales depends on the amount of training data, but rationales tend to hurt robustness in higher-resource settings (7.6K training examples) for these models. 

These general trends are similar to those for \nli, with more improvements from self-rationalization in lower-resource settings and some degradation in higher-resource settings. However, unlike for \nli, the results are not always monotonic in the amount of training data, particularly for \bartbase and \gptmedium on \deltaspread. In addition, for \gptlarge, results on \deltaspread improve with increasing data size, opposite to the general trend. Furthermore, improvements in \devhard are similar to standard errors, except for \tfivebase and n=1K, suggesting that even in low-resource settings, self-rationalization does not notably improve robustness for \cqa.

{The varied results for \cqa and lack of consistency between \nli and \cqa may be influenced by the differing numbers of artifacts in the datasets; in particular, perhaps self-rationalization training has a larger effect on robustness to spurious correlations when there are more spurious correlations in the training data (as in \snli but not \ecqa). We leave it to future work to investigate the impact of artifacts in training data on effect of rationales. The differences between \nli and \cqa also suggest that evaluations solely based on \nli may not cleanly transfer to other tasks; this finding provides further evidence that the benefits of rationales are task-dependent \citep{carton-etal-2020-evaluating, Palaskar2022OnAI} and that evaluations on one task such as \nli alone are not comprehensive enough to draw general conclusions about the utility of rationales.}

\begin{figure}[htb]
    \centering
    \includegraphics[height=0.23\textheight]{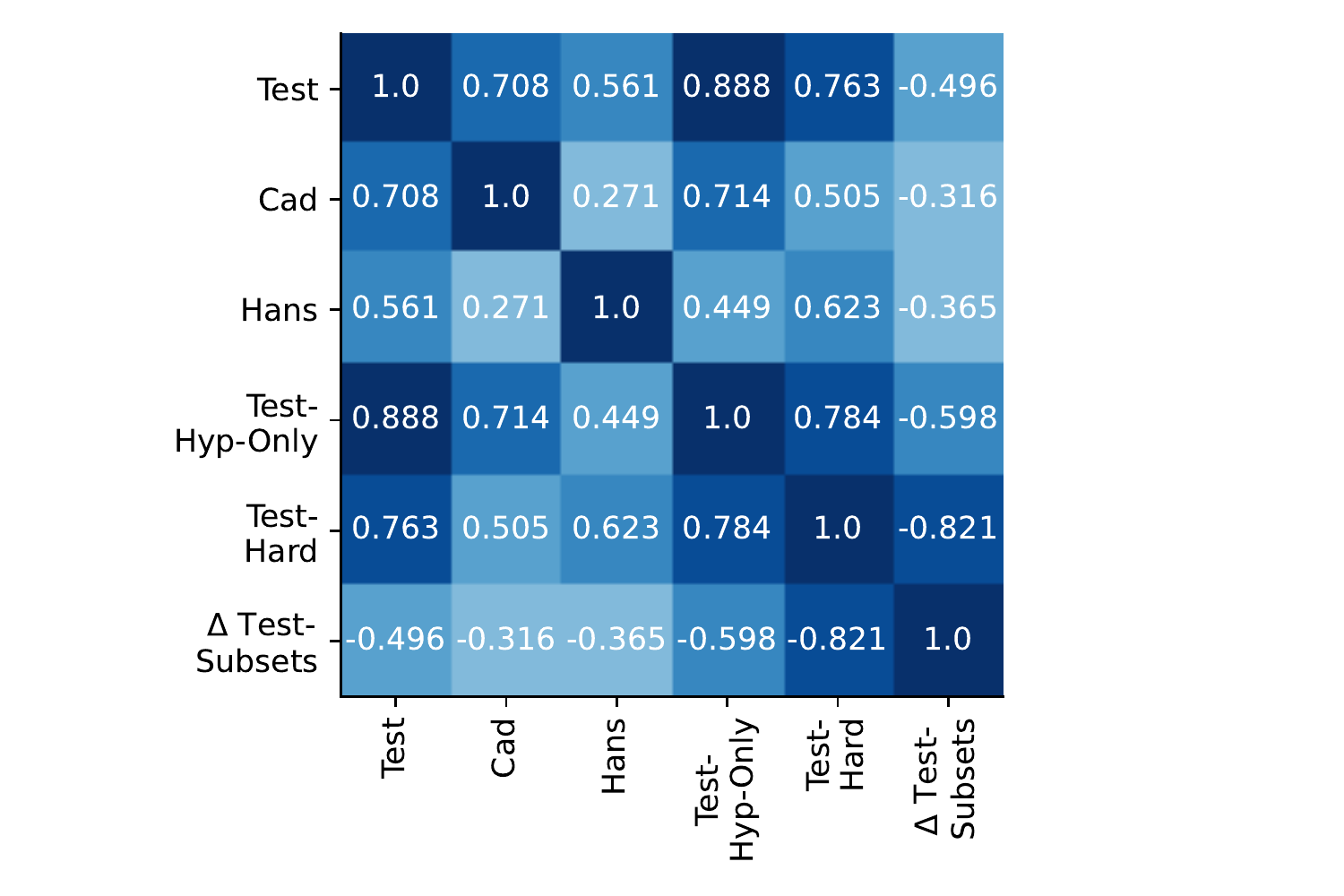}
    \vspace{-4mm}
\caption{Pearson correlations between results on pairs of evaluation metrics. Each cell color is determined by absolute value of the correlation coefficient.}\label{fig:correlations}
\end{figure}

\begin{table*}[t]\footnotesize\centering\begin{tabular}{lrrrrrr}\toprule
& \multicolumn{1}{c}{\dev}& \multicolumn{1}{c}{\cad}& \multicolumn{1}{c}{\hans}& \multicolumn{1}{c}{\begin{tabular}{@{}c@{}}\textsc{Test} \\ \textsc{Hyp}\end{tabular}} & \multicolumn{1}{c}{\begin{tabular}{@{}c@{}}\textsc{Test} \\ \textsc{Hard}\end{tabular}}& \multicolumn{1}{c}{\begin{tabular}{@{}c@{}}$\Delta$ \textsc{Test} \\ \textsc{Subsets}\end{tabular}} \\
& \multicolumn{1}{c}{$\uparrow$} & \multicolumn{1}{c}{$\uparrow$} & \multicolumn{1}{c}{$\uparrow$} & \multicolumn{1}{c}{$\uparrow$} & \multicolumn{1}{c}{$\uparrow$} & \multicolumn{1}{c}{$\downarrow$} \\
\cmidrule(lr){2-7}
\textbf{1K}\\



no rationales & 84.62~~~$\pm$0.31 & 71.20~~~$\pm$0.27 & 50.18~~~$\pm$0.08 & 73.68~~~$\pm$0.40 & 73.82~~~$\pm$0.85 & 14.70~~~$\pm$1.50 \\

original & \cellcolor{blue! 7.3500000000000005} +1.05~~~$\pm$0.21 & \cellcolor{blue! 12.25} +1.75~~~$\pm$0.34 & \cellcolor{blue! 3.08} +0.44~~~$\pm$0.19 & \cellcolor{blue! 10.92} +1.56~~~$\pm$0.40 & \cellcolor{blue! 14.280000000000001} +2.04~~~$\pm$0.91 & \cellcolor{blue! 14.280000000000001} -2.04~~~$\pm$1.93 \\

shuffled & \cellcolor{red! 4.13} -0.59~~~$\pm$0.37 & \cellcolor{red! 14.77} -2.11~~~$\pm$0.45 & \cellcolor{blue! 6.02} +0.86~~~$\pm$0.52 & \cellcolor{red! 12.67} -1.81~~~$\pm$0.50 & \cellcolor{red! 17.57} -2.51~~~$\pm$1.60 & \cellcolor{red! 29.05} +4.15~~~$\pm$2.85 \\
\bottomrule\end{tabular}\caption{Comparison between training \bartlarge on 1K training instances with original rationales in \esnli vs. shuffled rationales across 5 random seeds. We report means, as well as standard errors of the means.
}\label{tab:shuffled}\end{table*}

\subsection{Effect of Model Size}
\label{ssec:model_size}

For \nli, for the \gpt and \bart models, we find that increasing model size leads to increasing gains in robustness: Self-rationalization leads to larger improvements in robustness for \bartlarge than for \bartbase, and similarly for \gptlarge and \gptmedium (except for when $n$=2.5K); furthermore, we do not observe the same degradation in robustness for \bartlarge in higher-resource settings that we observe for \bartbase. For the \tfive models, self-rationalization generally leads to less degradation in robustness for \tfivelarge than for \tfivebase. For \cqa, we observe a similar trend: self-rationalization generally leads to larger improvements in robustness for \bartlarge than for \bartbase, for \gptlarge than for \gptmedium, and for \tfivelarge than for \tfivebase (except for when $n$=1K).

Thus, our results suggest that, within model families, increasing model size may improve effects on robustness from training with rationales. {Previous work has shown that rationales improve in-domain performance only for larger models, in both fine-tuning \citep{scratchpads} and in-context learning \citep{chain-of-thought, deepmind-prompt}; our results can be seen as an extension of this finding to the effects of training with rationales on \emph{robustness}. It is worth noting that the trends we observe appear to be specific to model families, \ie increasing model size has no noticeable effect when not conditioning on model family.}

\subsection{Correlation between robustness metrics}
\label{ssec:correlation}

To determine how results on different robustness metrics relate to each other, we compute their correlations. {These correlations should indicate how much insight we can get into the overall impact of self-rationalization on a model's robustness by only looking at select metrics.
For each pair of metrics in Figure~\ref{fig:nli-freeform}, we aggregate the differences in performance between baseline and self-rationalization performance on those metrics in all evaluation settings (\eg model type, training data size), and compute the Pearson Correlation of these values. 

As shown in Figure~\ref{fig:correlations}, results on the ``hard'' subsets of original test data (\devhard \& \snlihard) are overall correlated with the results on out-of-domain challenge datasets; the lowest correlation we observe for these subsets is between \snlihard and \hans, with Pearson coefficient 0.449. Furthermore, \cad and \hans, the manually annotated challenge sets, show low correlation with each other, with a Pearson coefficient of 0.271, {suggesting that out-of-domain performance does not straightforwardly reflect all aspects of robustness.}
We also observe that in-domain test performance is not always highly correlated with robustness metrics, with Pearson coefficient magnitudes as low as 0.496; this result suggests that difference in test performance is not entirely predictive of the effect of self-rationalization on robustness. In other words, training with rationales has effects on robustness that go beyond facilitating or hurting in-domain task performance.

\subsection{Effect of rationale content}
\label{ssec:rationale_content}

\paragraph{Shuffled explanations}
{One hypothesis for why training models to output rationales in addition to predictions may improve robustness is that it serves as a form of regularization; under this hypothesis, training to output even rationales with low explanatory power might improve robustness to spurious correlations by reducing overfitting.}

To determine to what extent rationale \emph{content} influences effects on robustness, we experiment with shuffling rationales during training such that the rationale for a given input no longer explains that input. 
Results from training \bartlarge with shuffled rationales for \nli are shown in Table~\ref{tab:shuffled}. {We also report results for \bartbase, \gptmedium, and \tfivelarge, which follow a similar trend, in Table~\ref{tab:all-shuffled} in the Appendix.}
We find that, as expected, training with shuffled rationales leads to worse robustness compared to training with original rationales, except on \hans.

\paragraph{Different \ecqa rationales}
We also experiment with training \bartlarge with the different rationale types in the \ecqa dataset, depicted in Table~\ref{tab:input-output}. {Results for \bartbase, \gptmedium, and \tfivelarge, which follow a similar trend, are shown in Table~\ref{tab:all-ecqa} in the Appendix.}

``Positive'' rationales explain why the gold answer is correct for a given question, ``negative'' rationales explain why other choices are incorrect, and ``freeflow'' rationales combine positive and negative rationales into a coherent and free-flowing paragraph and thus constitute freeform contrastive rationales.\footnote{{\emph{Contrastive} explanations explain why answers are correct compared to alternative (incorrect) answers \citep{Miller2019ExplanationIA, ross-etal-2021-explaining, jacovi-etal-2021-contrastive}.}}
As shown in Table~\ref{tab:cqa_freeform}, training with 1K positive rationales improves performance on both \dev \& \devhard and decreases \deltaspread. In contrast, training with 1K negative or freeflow rationales hurts performance on \dev \& \devhard. We also observe that training with freeflow rationales generally leads to worse results than positive rationales and better results than negative rationales. In contrast to prior findings on the benefits of contrastive rationales \citep{paranjape-etal-2021-prompting, schuster-etal-2021-get}, our results suggest that contrastive rationales do not always provide more learning benefits than non-contrastive rationales, given that training with freeflow rationales hurts robustness compare to the non-contrastive positive rationales.

A possible explanation for the differences in effects from training with these different rationale types is their varying lengths. As shown in Table~\ref{tab:input-output}, negative and freeflow rationales are longer than positive rationales.\footnote{The mean number of rationale tokens are 60, 62, and 29 for negative, freeflow, and positive rationales respectively.} To rule out this explanation, we also train \bartlarge with length-controlled negative and freeflow rationales, in which we truncate their lengths to 96 tokens, the maximum length used to train with positive rationales. As shown in Table~\ref{tab:cqa_freeform}, we still observe degradation in both task performance and robustness when using negative or freeflow rationales rather than positive rationales. These consistent results suggest that rationale content, rather than length, indeed influences learning. 

Another possible explanation 
for these varied effects is that
the topical relevance of rationales to gold labels may influence their utility in training. Positive rationales, as explanations of gold answers, are more topically related to gold answers than negative rationales, while freeflow rationales have topical relevance between those of positive and negative rationales. 
We observe that the effects of training with these rationale types align with their levels of topical relevance. Future work can further explore how properties like topical relevance influence the utility of rationales.

\begin{table}[t]\footnotesize\centering\begin{tabular}{lccc}\toprule
& \multicolumn{1}{c}{\dev}& \multicolumn{1}{c}{\begin{tabular}{@{}c@{}}\textsc{Test} \\ \textsc{Hard}\end{tabular}} & \begin{tabular}{@{}c@{}}$\Delta$ \textsc{Test} \\ \textsc{Subsets}\end{tabular}\\
& $\uparrow$ & $\uparrow$ & $\downarrow$\\\cmidrule(lr){2-2}\cmidrule(lr){3-3} \cmidrule(lr){4-4}
no rationales & 48.3~~~$\pm$0.8 & 49.0~~~$\pm$1.0 & 19.9~~~$\pm$1.8 \\\midrule
positive & \cellcolor{blue! 15.400000000000002} +2.2~~~$\pm$1.3 & \cellcolor{blue! 15.400000000000002} +2.2~~~$\pm$1.0 & \cellcolor{blue! 13.299999999999999} -1.9~~~$\pm$3.0 \\\midrule
 freeflow & \cellcolor{red! 16.099999999999998} -2.3~~~$\pm$1.1 & \cellcolor{red! 11.9} -1.7~~~$\pm$1.2 & \cellcolor{blue! 16.8} -2.4~~~$\pm$2.5 \\
 freeflow* & \cellcolor{red! 23.099999999999998} -3.3~~~$\pm$0.4 & \cellcolor{red! 7.700000000000001} -1.1~~~$\pm$2.7 & \cellcolor{red! 15.400000000000002} +2.2~~~$\pm$4.3 \\\midrule
 negative & \cellcolor{red! 28.699999999999996} -4.1~~~$\pm$1.2 & \cellcolor{red! 25.900000000000002} -3.7~~~$\pm$1.8 & \cellcolor{blue! 9.799999999999999} -1.4~~~$\pm$2.1 \\
negative* & \cellcolor{red! 37.1} -5.3~~~$\pm$0.3 & \cellcolor{red! 11.9} -1.7~~~$\pm$0.5 & \cellcolor{blue! 11.200000000000001} -1.6~~~$\pm$2.0 \\

\bottomrule\end{tabular}\caption{Effect of training \bartlarge with different types of rationales in \ecqa. Blue/red cells indicate improvement/worsening in performance compared to the baseline (no rationales, row 1).
* indicates length-controlled rationales, \ie truncation of the {negative} and {freeform} rationales to have the same length as the {positive} rationales. We report means across 5 random seeds, as well as standard errors of the means.}\label{tab:cqa_freeform}\end{table}

\section{Conclusion}

We investigate to what extent training models to rationalize their predictions affects their robustness to spurious correlations. We experiment with encoder-decoder and decoder-only models ranging in size from 140 to 774 million parameters across two tasks---natural language inference and commonsense question-answering---and measure reliance on spurious correlations through both manually-annotated, out-of-domain challenge sets and challenging in-domain subsets of original test sets. We find that the effects of self-rationalization are model- and task-specific: While self-rationalization can improve robustness to spurious correlations in lower-resource settings for some models and tasks, it tends to exacerbate reliance on spurious correlations in higher-resource settings. Furthermore, larger models tend to benefit more from rationales, and rationale content influences rationale utility in improving robustness.

{The variability of our results suggests that, despite the appeal of self-rationalization models for increasing model trustworthiness by facilitating debugging and interaction with end-users {\citep{formalizing-trust}}, training models to self-rationalize can have the unintended effect of increasing reliance on spurious features and biases, thereby decreasing the models' trustworthiness. Thus, appropriate care should be taken when training self-rationalization models with the goal of creating trustworthy models. Future work can investigate how to alleviate these harms while retaining the interpretability benefits of models that can rationalize their predictions.} 

\section{Limitations}

Conducting the analysis in this work required training over 700 models, particularly because the variability of model robustness requires training multiple models, governed by different random seeds, for every evaluation setting of interest. Thus, a main limitation of replicating this work is its computational demand.

Furthermore, even with the scale of our experiments, we do not exhaustively experiment with all possible evaluation settings of interest. Most notably, we focus our analysis on a standard way of training self-rationalization models---training generation models end-to-end to output rationales after their predictions; future work can investigate how our findings translate to other methods for training with rationales. \newhighlight{In addition, while many evaluation sets targeting robustness exist for \nli, they do not for \cqa; thus, our evaluation of robustness to spurious correlations for \cqa were limited. Future work can develop more tests for evaluating robustness for tasks beyond \nli.}

\section*{Acknowledgements}
We thank Howard Chen, Chenhao Tan, Joe Stacey, Marek Rei, members of the AllenNLP team, and anonymous reviewers for their helpful feedback.

\bibliography{anthology,custom}

\begin{thebibliography}{44}
\expandafter\ifx\csname natexlab\endcsname\relax\def\natexlab#1{#1}\fi

\bibitem[{Aggarwal et~al.(2021)Aggarwal, Mandowara, Agrawal, Khandelwal,
  Singla, and Garg}]{ecqa}
Shourya Aggarwal, Divyanshu Mandowara, Vishwajeet Agrawal, Dinesh Khandelwal,
  Parag Singla, and Dinesh Garg. 2021.
\newblock \href {https://doi.org/10.18653/v1/2021.acl-long.238} {{E}xplanations
  for {C}ommonsense{QA}: {N}ew {D}ataset and {M}odels}.
\newblock In \emph{Proceedings of the 59th Annual Meeting of the Association
  for Computational Linguistics and the 11th International Joint Conference on
  Natural Language Processing (Volume 1: Long Papers)}, pages 3050--3065,
  Online. Association for Computational Linguistics.

\bibitem[{Alvarez{-}Melis and Jaakkola(2018)}]{melis-self-explaining-robust}
David Alvarez{-}Melis and Tommi~S. Jaakkola. 2018.
\newblock \href
  {https://proceedings.neurips.cc/paper/2018/file/3e9f0fc9b2f89e043bc6233994dfcf76-Paper.pdf}
  {Towards robust interpretability with self-explaining neural networks}.
\newblock \emph{NeurIPS}.

\bibitem[{Bhat et~al.(2021)Bhat, Sordoni, and Mukherjee}]{bhat-etal-2021-self}
Meghana~Moorthy Bhat, Alessandro Sordoni, and Subhabrata Mukherjee. 2021.
\newblock \href {https://doi.org/10.18653/v1/2021.emnlp-main.836}
  {Self-training with few-shot rationalization}.
\newblock In \emph{Proceedings of the 2021 Conference on Empirical Methods in
  Natural Language Processing}, pages 10702--10712, Online and Punta Cana,
  Dominican Republic. Association for Computational Linguistics.

\bibitem[{Bowman et~al.(2015)Bowman, Angeli, Potts, and
  Manning}]{bowman-etal-2015-large}
Samuel~R. Bowman, Gabor Angeli, Christopher Potts, and Christopher~D. Manning.
  2015.
\newblock \href {https://doi.org/10.18653/v1/D15-1075} {A large annotated
  corpus for learning natural language inference}.
\newblock In \emph{Proceedings of the 2015 Conference on Empirical Methods in
  Natural Language Processing}, pages 632--642, Lisbon, Portugal. Association
  for Computational Linguistics.

\bibitem[{Camburu et~al.(2018)Camburu, Rockt\"{a}schel, Lukasiewicz, and
  Blunsom}]{esnli}
Oana-Maria Camburu, Tim Rockt\"{a}schel, Thomas Lukasiewicz, and Phil Blunsom.
  2018.
\newblock \href
  {http://papers.nips.cc/paper/8163-e-snli-natural-language-inference-with-natural-language-explanations.pdf}
  {e-snli: Natural language inference with natural language explanations}.
\newblock In S.~Bengio, H.~Wallach, H.~Larochelle, K.~Grauman, N.~Cesa-Bianchi,
  and R.~Garnett, editors, \emph{Advances in Neural Information Processing
  Systems 31}, pages 9539--9549. Curran Associates, Inc.

\bibitem[{Carton et~al.(2020)Carton, Rathore, and
  Tan}]{carton-etal-2020-evaluating}
Samuel Carton, Anirudh Rathore, and Chenhao Tan. 2020.
\newblock \href {https://doi.org/10.18653/v1/2020.emnlp-main.747} {Evaluating
  and characterizing human rationales}.
\newblock In \emph{Proceedings of the 2020 Conference on Empirical Methods in
  Natural Language Processing (EMNLP)}, pages 9294--9307, Online. Association
  for Computational Linguistics.

\bibitem[{Chen et~al.(2022)Chen, He, Narasimhan, and
  Chen}]{rationalization-robustness}
Howard Chen, Jacqueline He, Karthik Narasimhan, and Danqi Chen. 2022.
\newblock \href {https://arxiv.org/abs/2204.11790} {Can rationalization improve
  robustness?}

\bibitem[{DeYoung et~al.(2020)DeYoung, Jain, Rajani, Lehman, Xiong, Socher, and
  Wallace}]{deyoung-etal-2020-eraser}
Jay DeYoung, Sarthak Jain, Nazneen~Fatema Rajani, Eric Lehman, Caiming Xiong,
  Richard Socher, and Byron~C. Wallace. 2020.
\newblock \href {https://doi.org/10.18653/v1/2020.acl-main.408} {{ERASER}: {A}
  benchmark to evaluate rationalized {NLP} models}.
\newblock In \emph{Proceedings of the 58th Annual Meeting of the Association
  for Computational Linguistics}, pages 4443--4458, Online. Association for
  Computational Linguistics.

\bibitem[{Gardner et~al.(2021)Gardner, Merrill, Dodge, Peters, Ross, Singh, and
  Smith}]{gardner-etal-2021-competency}
Matt Gardner, William Merrill, Jesse Dodge, Matthew Peters, Alexis Ross, Sameer
  Singh, and Noah~A. Smith. 2021.
\newblock \href {https://doi.org/10.18653/v1/2021.emnlp-main.135} {Competency
  problems: On finding and removing artifacts in language data}.
\newblock In \emph{Proceedings of the 2021 Conference on Empirical Methods in
  Natural Language Processing}, pages 1801--1813, Online and Punta Cana,
  Dominican Republic. Association for Computational Linguistics.

\bibitem[{Gururangan et~al.(2018)Gururangan, Swayamdipta, Levy, Schwartz,
  Bowman, and Smith}]{gururangan-etal-2018-annotation}
Suchin Gururangan, Swabha Swayamdipta, Omer Levy, Roy Schwartz, Samuel Bowman,
  and Noah~A. Smith. 2018.
\newblock \href {https://doi.org/10.18653/v1/N18-2017} {Annotation artifacts in
  natural language inference data}.
\newblock In \emph{Proceedings of the 2018 Conference of the North {A}merican
  Chapter of the Association for Computational Linguistics: Human Language
  Technologies, Volume 2 (Short Papers)}, pages 107--112, New Orleans,
  Louisiana. Association for Computational Linguistics.

\bibitem[{Hancock et~al.(2018)Hancock, Varma, Wang, Bringmann, Liang, and
  R{\'e}}]{hancock-etal-2018-training}
Braden Hancock, Paroma Varma, Stephanie Wang, Martin Bringmann, Percy Liang,
  and Christopher R{\'e}. 2018.
\newblock \href {https://doi.org/10.18653/v1/P18-1175} {Training classifiers
  with natural language explanations}.
\newblock In \emph{Proceedings of the 56th Annual Meeting of the Association
  for Computational Linguistics (Volume 1: Long Papers)}, pages 1884--1895,
  Melbourne, Australia. Association for Computational Linguistics.

\bibitem[{Hase and Bansal(2021)}]{expl-formal-framework}
Peter Hase and Mohit Bansal. 2021.
\newblock \href {http://arxiv.org/abs/2102.02201} {When can models learn from
  explanations? {A} formal framework for understanding the roles of explanation
  data}.

\bibitem[{Jacovi et~al.(2021{\natexlab{a}})Jacovi, Marasovi\'{c}, Miller, and
  Goldberg}]{formalizing-trust}
Alon Jacovi, Ana Marasovi\'{c}, Tim Miller, and Yoav Goldberg.
  2021{\natexlab{a}}.
\newblock \href {https://doi.org/10.1145/3442188.3445923} {Formalizing trust in
  artificial intelligence: Prerequisites, causes and goals of human trust in
  ai}.
\newblock In \emph{Proceedings of the 2021 ACM Conference on Fairness,
  Accountability, and Transparency}, FAccT '21, page 624–635, New York, NY,
  USA. Association for Computing Machinery.

\bibitem[{Jacovi et~al.(2021{\natexlab{b}})Jacovi, Swayamdipta, Ravfogel,
  Elazar, Choi, and Goldberg}]{jacovi-etal-2021-contrastive}
Alon Jacovi, Swabha Swayamdipta, Shauli Ravfogel, Yanai Elazar, Yejin Choi, and
  Yoav Goldberg. 2021{\natexlab{b}}.
\newblock \href {https://doi.org/10.18653/v1/2021.emnlp-main.120} {Contrastive
  explanations for model interpretability}.
\newblock In \emph{Proceedings of the 2021 Conference on Empirical Methods in
  Natural Language Processing}, pages 1597--1611, Online and Punta Cana,
  Dominican Republic. Association for Computational Linguistics.

\bibitem[{Kaushik et~al.(2021)Kaushik, Setlur, Hovy, and
  Lipton}]{kaushik2021learning}
Divyansh Kaushik, Amrith Setlur, Eduard Hovy, and Zachary~C Lipton. 2021.
\newblock \href {https://arxiv.org/abs/2010.02114} {Explaining the efficacy of
  counterfactually augmented data}.
\newblock \emph{International Conference on Learning Representations (ICLR)}.

\bibitem[{Lampinen et~al.(2022)Lampinen, Dasgupta, Chan, Matthewson, Tessler,
  Creswell, McClelland, Wang, and Hill}]{deepmind-prompt}
Andrew~K. Lampinen, Ishita Dasgupta, Stephanie C.~Y. Chan, Kory Matthewson,
  Michael~Henry Tessler, Antonia Creswell, James~L. McClelland, Jane~X. Wang,
  and Felix Hill. 2022.
\newblock \href {https://arxiv.org/abs/2204.02329} {Can language models learn
  from explanations in context?}

\bibitem[{Lester et~al.(2021)Lester, Al-Rfou, and
  Constant}]{lester-etal-2021-power}
Brian Lester, Rami Al-Rfou, and Noah Constant. 2021.
\newblock \href {https://doi.org/10.18653/v1/2021.emnlp-main.243} {The power of
  scale for parameter-efficient prompt tuning}.
\newblock In \emph{Proceedings of the 2021 Conference on Empirical Methods in
  Natural Language Processing}, pages 3045--3059, Online and Punta Cana,
  Dominican Republic. Association for Computational Linguistics.

\bibitem[{Lewis et~al.(2020)Lewis, Liu, Goyal, Ghazvininejad, Mohamed, Levy,
  Stoyanov, and Zettlemoyer}]{lewis-etal-2020-bart}
Mike Lewis, Yinhan Liu, Naman Goyal, Marjan Ghazvininejad, Abdelrahman Mohamed,
  Omer Levy, Veselin Stoyanov, and Luke Zettlemoyer. 2020.
\newblock \href {https://doi.org/10.18653/v1/2020.acl-main.703} {{BART}:
  Denoising sequence-to-sequence pre-training for natural language generation,
  translation, and comprehension}.
\newblock In \emph{Proceedings of the 58th Annual Meeting of the Association
  for Computational Linguistics}, pages 7871--7880, Online. Association for
  Computational Linguistics.

\bibitem[{Lombrozo(2016)}]{explanation-lombrozo}
Tania Lombrozo. 2016.
\newblock \href {https://doi.org/10.1016/j.tics.2016.08.001} {Explanatory
  preferences shape learning and inference}.
\newblock \emph{Trends in Cognitive Sciences}, 20(10):748--759.

\bibitem[{McCoy et~al.(2019)McCoy, Pavlick, and Linzen}]{mccoy-etal-2019-right}
Tom McCoy, Ellie Pavlick, and Tal Linzen. 2019.
\newblock \href {https://doi.org/10.18653/v1/P19-1334} {Right for the wrong
  reasons: Diagnosing syntactic heuristics in natural language inference}.
\newblock In \emph{Proceedings of the 57th Annual Meeting of the Association
  for Computational Linguistics}, pages 3428--3448, Florence, Italy.
  Association for Computational Linguistics.

\bibitem[{Miller(2019)}]{Miller2019ExplanationIA}
Tim Miller. 2019.
\newblock \href
  {https://www.sciencedirect.com/science/article/abs/pii/S0004370218305988}
  {Explanation in {A}rtificial {I}ntelligence: {Insights} from the social
  sciences}.
\newblock \emph{Artificial Intelligence}, 267:1--38.

\bibitem[{Narang et~al.(2020)Narang, Raffel, Lee, Roberts, Fiedel, and
  Malkan}]{wt5}
Sharan Narang, Colin Raffel, Katherine Lee, Adam Roberts, Noah Fiedel, and
  Karishma Malkan. 2020.
\newblock \href {https://arxiv.org/abs/2004.14546} {Wt5?! training text-to-text
  models to explain their predictions}.

\bibitem[{Nye et~al.(2021)Nye, Andreassen, Gur{-}Ari, Michalewski, Austin,
  Bieber, Dohan, Lewkowycz, Bosma, Luan, Sutton, and Odena}]{scratchpads}
Maxwell~I. Nye, Anders~Johan Andreassen, Guy Gur{-}Ari, Henryk Michalewski,
  Jacob Austin, David Bieber, David Dohan, Aitor Lewkowycz, Maarten Bosma,
  David Luan, Charles Sutton, and Augustus Odena. 2021.
\newblock \href {https://arxiv.org/abs/2112.00114} {Show your work: Scratchpads
  for intermediate computation with language models}.

\bibitem[{Palaskar et~al.(2022)Palaskar, Bhagia, Bisk, Metze, Black, and
  Marasovi{\'c}}]{Palaskar2022OnAI}
Shruti Palaskar, Akshita Bhagia, Yonatan Bisk, Florian Metze, Alan~W. Black,
  and Ana Marasovi{\'c}. 2022.
\newblock \href {https://arxiv.org/abs/2205.11686} {On advances in text
  generation from images beyond captioning: A case study in
  self-rationalization}.
\newblock \emph{Proceedings of the 2022 Conference on Empirical Methods in
  Natural Language Processing (EMNLP)}.

\bibitem[{Paranjape et~al.(2021)Paranjape, Michael, Ghazvininejad, Hajishirzi,
  and Zettlemoyer}]{paranjape-etal-2021-prompting}
Bhargavi Paranjape, Julian Michael, Marjan Ghazvininejad, Hannaneh Hajishirzi,
  and Luke Zettlemoyer. 2021.
\newblock \href {https://doi.org/10.18653/v1/2021.findings-acl.366} {Prompting
  contrastive explanations for commonsense reasoning tasks}.
\newblock In \emph{Findings of the Association for Computational Linguistics:
  ACL-IJCNLP 2021}, pages 4179--4192, Online. Association for Computational
  Linguistics.

\bibitem[{Poliak et~al.(2018)Poliak, Naradowsky, Haldar, Rudinger, and
  Van~Durme}]{poliak-etal-2018-hypothesis}
Adam Poliak, Jason Naradowsky, Aparajita Haldar, Rachel Rudinger, and Benjamin
  Van~Durme. 2018.
\newblock \href {https://doi.org/10.18653/v1/S18-2023} {Hypothesis only
  baselines in natural language inference}.
\newblock In \emph{Proceedings of the Seventh Joint Conference on Lexical and
  Computational Semantics}, pages 180--191, New Orleans, Louisiana. Association
  for Computational Linguistics.

\bibitem[{Pruthi et~al.(2022)Pruthi, Bansal, Dhingra, Soares, Collins, Lipton,
  Neubig, and Cohen}]{teacher-student-expl}
Danish Pruthi, Rachit Bansal, Bhuwan Dhingra, Livio~Baldini Soares, Michael
  Collins, Zachary~C. Lipton, Graham Neubig, and William~W. Cohen. 2022.
\newblock \href
  {https://direct.mit.edu/tacl/article/doi/10.1162/tacl_a_00465/110436/Evaluating-Explanations-How-Much-Do-Explanations}
  {{Evaluating Explanations: How Much Do Explanations from the Teacher Aid
  Students?}}
\newblock \emph{Transactions of the Association for Computational Linguistics},
  10:359--375.

\bibitem[{Radford et~al.(2019)Radford, Wu, Child, Luan, Amodei, and
  Sutskever}]{gpt}
Alec Radford, Jeff Wu, Rewon Child, David Luan, Dario Amodei, and Ilya
  Sutskever. 2019.
\newblock \href
  {https://d4mucfpksywv.cloudfront.net/better-language-models/language_models_are_unsupervised_multitask_learners.pdf}
  {Language models are unsupervised multitask learners}.

\bibitem[{Raffel et~al.(2020)Raffel, Shazeer, Roberts, Lee, Narang, Matena,
  Zhou, Li, and Liu}]{2020t5}
Colin Raffel, Noam Shazeer, Adam Roberts, Katherine Lee, Sharan Narang, Michael
  Matena, Yanqi Zhou, Wei Li, and Peter~J. Liu. 2020.
\newblock \href {http://jmlr.org/papers/v21/20-074.html} {Exploring the limits
  of transfer learning with a unified text-to-text transformer}.
\newblock \emph{Journal of Machine Learning Research}, 21(140):1--67.

\bibitem[{Rajani et~al.(2019)Rajani, McCann, Xiong, and
  Socher}]{rajani-etal-2019-explain}
Nazneen~Fatema Rajani, Bryan McCann, Caiming Xiong, and Richard Socher. 2019.
\newblock \href {https://doi.org/10.18653/v1/P19-1487} {Explain yourself!
  leveraging language models for commonsense reasoning}.
\newblock In \emph{Proceedings of the 57th Annual Meeting of the Association
  for Computational Linguistics}, pages 4932--4942, Florence, Italy.
  Association for Computational Linguistics.

\bibitem[{Ross et~al.(2021)Ross, Marasovi{\'c}, and
  Peters}]{ross-etal-2021-explaining}
Alexis Ross, Ana Marasovi{\'c}, and Matthew Peters. 2021.
\newblock \href {https://doi.org/10.18653/v1/2021.findings-acl.336} {Explaining
  {NLP} models via minimal contrastive editing ({M}i{CE})}.
\newblock In \emph{Findings of the Association for Computational Linguistics:
  ACL-IJCNLP 2021}, pages 3840--3852, Online. Association for Computational
  Linguistics.

\bibitem[{Schuff et~al.(2021)Schuff, Yang, Adel, and
  Vu}]{schuff-etal-2021-external}
Hendrik Schuff, Hsiu-Yu Yang, Heike Adel, and Ngoc~Thang Vu. 2021.
\newblock \href {https://doi.org/10.18653/v1/2021.blackboxnlp-1.3} {Does
  external knowledge help explainable natural language inference? automatic
  evaluation vs. human ratings}.
\newblock In \emph{Proceedings of the Fourth BlackboxNLP Workshop on Analyzing
  and Interpreting Neural Networks for NLP}, pages 26--41, Punta Cana,
  Dominican Republic. Association for Computational Linguistics.

\bibitem[{Schuster et~al.(2021)Schuster, Fisch, and
  Barzilay}]{schuster-etal-2021-get}
Tal Schuster, Adam Fisch, and Regina Barzilay. 2021.
\newblock \href {https://doi.org/10.18653/v1/2021.naacl-main.52} {Get your
  vitamin {C}! robust fact verification with contrastive evidence}.
\newblock In \emph{Proceedings of the 2021 Conference of the North American
  Chapter of the Association for Computational Linguistics: Human Language
  Technologies}, pages 624--643, Online. Association for Computational
  Linguistics.

\bibitem[{Stacey et~al.(2022)Stacey, Belinkov, and Rei}]{Stacey2021NaturalLI}
Joe Stacey, Yonatan Belinkov, and Marek Rei. 2022.
\newblock \href {https://arxiv.org/abs/2104.08142} {Supervising model attention
  with human explanations for robust natural language inference}.
\newblock \emph{AAAI}, abs/2104.08142.

\bibitem[{Talmor et~al.(2019)Talmor, Herzig, Lourie, and
  Berant}]{talmor-etal-2019-commonsenseqa}
Alon Talmor, Jonathan Herzig, Nicholas Lourie, and Jonathan Berant. 2019.
\newblock \href {https://doi.org/10.18653/v1/N19-1421} {{C}ommonsense{QA}: A
  question answering challenge targeting commonsense knowledge}.
\newblock In \emph{Proceedings of the 2019 Conference of the North {A}merican
  Chapter of the Association for Computational Linguistics: Human Language
  Technologies, Volume 1 (Long and Short Papers)}, pages 4149--4158,
  Minneapolis, Minnesota. Association for Computational Linguistics.

\bibitem[{Utama et~al.(2021)Utama, Moosavi, Sanh, and
  Gurevych}]{utama-etal-2021-avoiding}
Prasetya Utama, Nafise~Sadat Moosavi, Victor Sanh, and Iryna Gurevych. 2021.
\newblock \href {https://doi.org/10.18653/v1/2021.emnlp-main.713} {Avoiding
  inference heuristics in few-shot prompt-based finetuning}.
\newblock In \emph{Proceedings of the 2021 Conference on Empirical Methods in
  Natural Language Processing}, pages 9063--9074, Online and Punta Cana,
  Dominican Republic. Association for Computational Linguistics.

\bibitem[{Wallace et~al.(2019)Wallace, Feng, Kandpal, Gardner, and
  Singh}]{wallace-etal-2019-universal}
Eric Wallace, Shi Feng, Nikhil Kandpal, Matt Gardner, and Sameer Singh. 2019.
\newblock \href {https://doi.org/10.18653/v1/D19-1221} {Universal adversarial
  triggers for attacking and analyzing {NLP}}.
\newblock In \emph{Proceedings of the 2019 Conference on Empirical Methods in
  Natural Language Processing and the 9th International Joint Conference on
  Natural Language Processing (EMNLP-IJCNLP)}, pages 2153--2162, Hong Kong,
  China. Association for Computational Linguistics.

\bibitem[{Wei et~al.(2022)Wei, Wang, Schuurmans, Bosma, Chi, Le, and
  Zhou}]{chain-of-thought}
Jason Wei, Xuezhi Wang, Dale Schuurmans, Maarten Bosma, Ed~Chi, Quoc Le, and
  Denny Zhou. 2022.
\newblock \href {https://arxiv.org/abs/2201.11903} {Chain of thought prompting
  elicits reasoning in large language models}.

\bibitem[{Wiegreffe et~al.(2021)Wiegreffe, Marasovi{\'c}, and
  Smith}]{wiegreffe-etal-2021-measuring}
Sarah Wiegreffe, Ana Marasovi{\'c}, and Noah~A. Smith. 2021.
\newblock \href {https://doi.org/10.18653/v1/2021.emnlp-main.804} {{M}easuring
  association between labels and free-text rationales}.
\newblock In \emph{Proceedings of the 2021 Conference on Empirical Methods in
  Natural Language Processing}, pages 10266--10284, Online and Punta Cana,
  Dominican Republic. Association for Computational Linguistics.

\bibitem[{Wu et~al.(2022)Wu, Gardner, Stenetorp, and
  Dasigi}]{wu-etal-2022-generating}
Yuxiang Wu, Matt Gardner, Pontus Stenetorp, and Pradeep Dasigi. 2022.
\newblock \href {https://aclanthology.org/2022.acl-long.190} {Generating data
  to mitigate spurious correlations in natural language inference datasets}.
\newblock In \emph{Proceedings of the 60th Annual Meeting of the Association
  for Computational Linguistics (Volume 1: Long Papers)}, pages 2660--2676,
  Dublin, Ireland. Association for Computational Linguistics.

\bibitem[{Ye and Durrett(2022)}]{unreliability-few-shot}
Xi~Ye and Greg Durrett. 2022.
\newblock \href {https://arxiv.org/abs/2205.03401} {The unreliability of
  explanations in few-shot in-context learning}.

\bibitem[{Zaidan et~al.(2007)Zaidan, Eisner, and
  Piatko}]{zaidan-etal-2007-using}
Omar Zaidan, Jason Eisner, and Christine Piatko. 2007.
\newblock \href {https://aclanthology.org/N07-1033} {Using {``}annotator
  rationales{''} to improve machine learning for text categorization}.
\newblock In \emph{Human Language Technologies 2007: The Conference of the
  North {A}merican Chapter of the Association for Computational Linguistics;
  Proceedings of the Main Conference}, pages 260--267, Rochester, New York.
  Association for Computational Linguistics.

\bibitem[{Zelikman et~al.(2022)Zelikman, Wu, and Goodman}]{star}
Eric Zelikman, Yuhuai Wu, and Noah~D. Goodman. 2022.
\newblock \href {https://arxiv.org/abs/2203.14465} {Star: Bootstrapping
  reasoning with reasoning}.

\bibitem[{Zhao and Vydiswaran(2021)}]{lirex}
Xinyan Zhao and V.~G.~Vinod Vydiswaran. 2021.
\newblock \href {https://arxiv.org/abs/2012.09157} {Lirex: Augmenting language
  inference with relevant explanation}.
\newblock \emph{AAAI}, abs/2012.09157.

\end{thebibliography}
\bibliographystyle{acl_natbib}

\appendix 

\begin{table*}[t]\footnotesize\centering\begin{tabular}{llllllll}\toprule
& \multicolumn{1}{c}{\dev}& \multicolumn{1}{c}{\cad}& \multicolumn{1}{c}{\hans}& \multicolumn{1}{c}{\begin{tabular}{@{}c@{}}\textsc{Test} \\ \textsc{Hyp}\end{tabular}} & \multicolumn{1}{c}{\begin{tabular}{@{}c@{}}\textsc{Test} \\ \textsc{Hard}\end{tabular}}& \multicolumn{1}{c}{\begin{tabular}{@{}c@{}}$\Delta$ \textsc{Test} \\ \textsc{Subsets}\end{tabular}} \\
& \multicolumn{1}{c}{$\uparrow$} & \multicolumn{1}{c}{$\uparrow$} & \multicolumn{1}{c}{$\uparrow$} & \multicolumn{1}{c}{$\uparrow$} & \multicolumn{1}{c}{$\uparrow$} & \multicolumn{1}{c}{$\downarrow$} \\
\cmidrule(lr){2-7}
\textbf{\bartbase}\\
no rationales & 75.37~~~$\pm$0.15 & 57.85~~~$\pm$0.61 & 51.30~~~$\pm$0.58 & 59.80~~~$\pm$0.67 & 56.86~~~$\pm$0.66 & 25.44~~~$\pm$2.75 \\
original & \cellcolor{blue! 1.47} +0.21~~~$\pm$0.18 & \cellcolor{blue! 7.28} +1.04~~~$\pm$0.63 & \cellcolor{blue! 6.72} +0.96~~~$\pm$0.62 & \cellcolor{red! 1.26} -0.18~~~$\pm$0.38 & \cellcolor{blue! 9.87} +1.41~~~$\pm$0.46 & \cellcolor{blue! 19.11} -2.73~~~$\pm$1.65 \\
shuffled & \cellcolor{red! 32.97} -4.71~~~$\pm$0.83 & \cellcolor{red! 39.199999999999996} -5.60~~~$\pm$1.11 & \cellcolor{red! 1.47} -0.21~~~$\pm$0.39 & \cellcolor{red! 39.199999999999996} -5.60~~~$\pm$1.05 & \cellcolor{red! 30.800000000000004} -4.40~~~$\pm$1.52 & \cellcolor{red! 14.700000000000001} +2.10~~~$\pm$3.81 \\\midrule
\textbf{\gptmedium}\\
no rationales & 66.39~~~$\pm$0.94 & 48.84~~~$\pm$0.81 & 52.66~~~$\pm$0.38 & 48.70~~~$\pm$1.20 & 47.59~~~$\pm$2.22 & 33.39~~~$\pm$2.19 \\
original & \cellcolor{blue! 7.840000000000001} +1.12~~~$\pm$0.66 & \cellcolor{blue! 5.6000000000000005} +0.80~~~$\pm$0.73 & \cellcolor{blue! 9.94} +1.42~~~$\pm$0.27 & \cellcolor{blue! 4.13} +0.59~~~$\pm$1.07 & \cellcolor{blue! 25.27} +3.61~~~$\pm$2.16 & \cellcolor{blue! 27.58} -3.94~~~$\pm$3.97 \\
shuffled & \cellcolor{red! 13.79} -1.97~~~$\pm$0.75 & \cellcolor{red! 17.150000000000002} -2.45~~~$\pm$0.95 & \cellcolor{blue! 6.72} +0.96~~~$\pm$0.55 & \cellcolor{red! 26.040000000000003} -3.72~~~$\pm$1.46 & \cellcolor{red! 0.0} 0.00~~~$\pm$3.09 & \cellcolor{red! 4.62} +0.66~~~$\pm$4.63 \\\midrule
\textbf{\bartlarge}\\
no rationales & 84.62~~~$\pm$0.31 & 71.20~~~$\pm$0.27 & 50.18~~~$\pm$0.08 & 73.68~~~$\pm$0.40 & 73.82~~~$\pm$0.85 & 14.70~~~$\pm$1.50 \\
original & \cellcolor{blue! 7.3500000000000005} +1.05~~~$\pm$0.21 & \cellcolor{blue! 12.25} +1.75~~~$\pm$0.34 & \cellcolor{blue! 3.08} +0.44~~~$\pm$0.19 & \cellcolor{blue! 10.92} +1.56~~~$\pm$0.40 & \cellcolor{blue! 14.280000000000001} +2.04~~~$\pm$0.91 & \cellcolor{blue! 14.280000000000001} -2.04~~~$\pm$1.93 \\
shuffled & \cellcolor{red! 4.13} -0.59~~~$\pm$0.37 & \cellcolor{red! 14.77} -2.11~~~$\pm$0.45 & \cellcolor{blue! 6.02} +0.86~~~$\pm$0.52 & \cellcolor{red! 12.67} -1.81~~~$\pm$0.50 & \cellcolor{red! 17.57} -2.51~~~$\pm$1.60 & \cellcolor{red! 29.050000000000004} +4.15~~~$\pm$2.86 \\\midrule
\textbf{\tfivelarge}\\
no rationales & 84.03~~~$\pm$0.34 & 71.08~~~$\pm$0.31 & 51.54~~~$\pm$0.31 & 74.26~~~$\pm$0.54 & 75.45~~~$\pm$0.71 & 10.46~~~$\pm$1.03 \\
original & \cellcolor{blue! 1.82} +0.26~~~$\pm$0.21 & \cellcolor{blue! 4.34} +0.62~~~$\pm$0.48 & \cellcolor{red! 7.909999999999999} -1.13~~~$\pm$0.17 & \cellcolor{red! 0.7000000000000001} -0.10~~~$\pm$0.40 & \cellcolor{red! 2.94} -0.42~~~$\pm$1.11 & \cellcolor{red! 14.42} +2.06~~~$\pm$2.33 \\
shuffled & \cellcolor{red! 4.69} -0.67~~~$\pm$0.41 & \cellcolor{red! 1.61} -0.23~~~$\pm$0.18 & \cellcolor{blue! 21.14} +3.02~~~$\pm$0.98 & \cellcolor{red! 5.25} -0.75~~~$\pm$0.70 & \cellcolor{red! 18.34} -2.62~~~$\pm$1.25 & \cellcolor{red! 22.89} +3.27~~~$\pm$2.46 \\
\bottomrule\end{tabular}\caption{Comparison between training with original rationales in \esnli vs. shuffled rationales with 1K instances for \nli. Blue/red cells indicate improvement/worsening in performance compared to the baseline (no rationales). We report means across 5 random seeds, as well as standard errors of the means.}\label{tab:all-shuffled}\end{table*}

\begin{table*}[t]\footnotesize\centering\begin{tabular}{lccc}\toprule
& \multicolumn{1}{c}{\dev}& \multicolumn{1}{c}{\begin{tabular}{@{}c@{}}\textsc{Test} \\ \textsc{Hard}\end{tabular}} & \begin{tabular}{@{}c@{}}$\Delta$ \textsc{Test} \\ \textsc{Subsets}\end{tabular}\\
& $\uparrow$ & $\uparrow$ & $\downarrow$\\\cmidrule(lr){2-2}\cmidrule(lr){3-3} \cmidrule(lr){4-4}
\textbf{\bartbase}\\
no rationales & 32.4~~~$\pm$0.3 & 33.6~~~$\pm$0.5 & 14.7~~~$\pm$1.4 \\\midrule
positive & \cellcolor{red! 0.0} 0.0~~~$\pm$0.4 & \cellcolor{blue! 9.799999999999999} +1.4~~~$\pm$0.8 & \cellcolor{blue! 20.3} -2.9~~~$\pm$2.3 \\\midrule
 freeflow & \cellcolor{red! 9.799999999999999} -1.4~~~$\pm$0.7 & \cellcolor{blue! 2.1} +0.3~~~$\pm$1.2 & \cellcolor{blue! 0.0} -0.0~~~$\pm$1.9 \\
  freeflow* & \cellcolor{red! 11.9} -1.7~~~$\pm$0.4 & \cellcolor{red! 4.2} -0.6~~~$\pm$0.8 & \cellcolor{blue! 0.0} -0.0~~~$\pm$3.1 \\\midrule
 negative & \cellcolor{red! 24.5} -3.5~~~$\pm$1.0 & \cellcolor{red! 17.5} -2.5~~~$\pm$1.2 & \cellcolor{blue! 14.0} -2.0~~~$\pm$1.6 \\
 negative* & \cellcolor{red! 21.7} -3.1~~~$\pm$0.8 & \cellcolor{red! 15.400000000000002} -2.2~~~$\pm$0.8 & \cellcolor{red! 26.599999999999998} +3.8~~~$\pm$1.7 \\\midrule\midrule
\textbf{\gptmedium}\\
no rationales & 30.6~~~$\pm$0.4 & 29.2~~~$\pm$1.5 & 13.3~~~$\pm$3.5 \\\midrule
positive & \cellcolor{blue! 16.099999999999998} +2.3~~~$\pm$0.3 & \cellcolor{blue! 6.3} +0.9~~~$\pm$1.5 & \cellcolor{red! 18.2} +2.6~~~$\pm$2.5 \\\midrule
 freeflow & \cellcolor{blue! 18.2} +2.6~~~$\pm$0.4 & \cellcolor{blue! 14.700000000000001} +2.1~~~$\pm$0.6 & \cellcolor{red! 18.900000000000002} +2.7~~~$\pm$2.8 \\
 freeflow* & \cellcolor{blue! 23.8} +3.4~~~$\pm$0.6 & \cellcolor{blue! 20.3} +2.9~~~$\pm$0.6 & \cellcolor{red! 12.6} +1.8~~~$\pm$2.4 \\\midrule
 negative & \cellcolor{blue! 6.3} +0.9~~~$\pm$0.2 & \cellcolor{blue! 5.6000000000000005} +0.8~~~$\pm$1.1 & \cellcolor{red! 12.6} +1.8~~~$\pm$3.7 \\
 negative* & \cellcolor{blue! 7.700000000000001} +1.1~~~$\pm$0.4 & \cellcolor{blue! 6.3} +0.9~~~$\pm$0.6 & \cellcolor{red! 9.1} +1.3~~~$\pm$1.9 \\\midrule\midrule
\textbf{\bartlarge}\\
no rationales & 48.3~~~$\pm$0.8 & 49.0~~~$\pm$1.0 & 19.9~~~$\pm$1.8 \\\midrule
positive & \cellcolor{blue! 15.400000000000002} +2.2~~~$\pm$1.3 & \cellcolor{blue! 15.400000000000002} +2.2~~~$\pm$1.0 & \cellcolor{blue! 13.299999999999999} -1.9~~~$\pm$3.0 \\\midrule
 freeflow & \cellcolor{red! 16.099999999999998} -2.3~~~$\pm$1.1 & \cellcolor{red! 11.9} -1.7~~~$\pm$1.2 & \cellcolor{blue! 16.8} -2.4~~~$\pm$2.5 \\
 freeflow* & \cellcolor{red! 23.099999999999998} -3.3~~~$\pm$0.4 & \cellcolor{red! 7.700000000000001} -1.1~~~$\pm$2.7 & \cellcolor{red! 15.400000000000002} +2.2~~~$\pm$4.3 \\\midrule
 negative & \cellcolor{red! 28.699999999999996} -4.1~~~$\pm$1.2 & \cellcolor{red! 25.900000000000002} -3.7~~~$\pm$1.8 & \cellcolor{blue! 9.799999999999999} -1.4~~~$\pm$2.1 \\
negative* & \cellcolor{red! 37.1} -5.3~~~$\pm$0.3 & \cellcolor{red! 11.9} -1.7~~~$\pm$0.5 & \cellcolor{blue! 11.200000000000001} -1.6~~~$\pm$2.0 \\\midrule\midrule
\textbf{\tfivelarge}\\
no rationales & 60.4~~~$\pm$0.7 & 61.5~~~$\pm$1.5 & 10.5~~~$\pm$1.9 \\\midrule
positive & \cellcolor{blue! 8.4} +1.2~~~$\pm$0.7 & \cellcolor{blue! 14.0} +2.0~~~$\pm$1.7 & \cellcolor{red! 16.8} +2.4~~~$\pm$1.3 \\\midrule
 freeflow & \cellcolor{blue! 7.0} +1.0~~~$\pm$0.9 & \cellcolor{blue! 11.200000000000001} +1.6~~~$\pm$1.3 & \cellcolor{red! 2.1} +0.3~~~$\pm$1.4 \\
freeflow* & \cellcolor{blue! 7.0} +1.0~~~$\pm$0.7 & \cellcolor{blue! 11.9} +1.7~~~$\pm$0.9 & \cellcolor{red! 14.700000000000001} +2.1~~~$\pm$1.2 \\\midrule
 negative & \cellcolor{red! 7.700000000000001} -1.1~~~$\pm$0.6 & \cellcolor{red! 9.1} -1.3~~~$\pm$0.8 & \cellcolor{red! 31.5} +4.5~~~$\pm$3.1 \\
 negative* & \cellcolor{red! 9.1} -1.3~~~$\pm$0.5 & \cellcolor{red! 11.200000000000001} -1.6~~~$\pm$0.7 & \cellcolor{red! 17.5} +2.5~~~$\pm$1.4 \\
\bottomrule\end{tabular}\caption{Effect of training with 1K instances using the different types of rationales in \ecqa. * indicates length-controlled rationales. Blue/red cells indicate improvement/worsening in performance compared to the baseline (no rationales, row 1). We report means across 5 random seeds, as well as standard errors of the means.
}\label{tab:all-ecqa}\end{table*}

\end{document}